%% file: iclr2027_conference.tex
\documentclass{article} 
\usepackage{iclr2027_conference,times}

\input{math_commands.tex}

\usepackage[utf8]{inputenc} 
\usepackage[T1]{fontenc}    
\usepackage{hyperref}       
\usepackage{url}            
\usepackage{booktabs}       
\usepackage{amsfonts}       
\usepackage{nicefrac}       
\usepackage{microtype}      
\usepackage{xcolor}         

\usepackage{orcidlink}

\usepackage{amsmath}
\usepackage{amssymb}
\usepackage{mathtools}

\usepackage{graphicx}
\usepackage{wrapfig}
\usepackage{booktabs}
\usepackage{multicol}
\usepackage{multirow}
\usepackage{colortbl}
\usepackage{caption}
\usepackage{makecell}
\usepackage{xspace}
\usepackage{cleveref}

\renewcommand{\paragraph}[1]{\vspace{2pt}\noindent\textbf{#1}}
\newcommand{\Method}{ClinSeekAgent\xspace}
\newcommand{\Benchmark}{ClinSeek-Bench\xspace}
\newcommand{\Baseline}{Curated Input\xspace}

\let\cite\citep

\usepackage{tabularx}
\usepackage{array}
\usepackage{seqsplit}
\usepackage{makecell}

\crefname{figure}{Fig.}{Fig.}
\crefname{table}{Tab.}{Tab.}
\crefname{section}{Sec.}{Sec.}
\crefname{appendix}{App.}{App.}

\title{ClinSeekAgent: Automating Multimodal Evidence Seeking for Agentic Clinical Reasoning}

\author{
\textbf{Juncheng Wu}$^{1*}$, \textbf{Letian Zhang}$^{1*}$, \textbf{Yuhan Wang}$^{1*}$, \textbf{Haoqin Tu}$^{1}$,
\textbf{Hardy Chen}$^{1}$, \textbf{Zijun Wang}$^{1}$, \\
\textbf{Cihang Xie}$^{1}$, \textbf{Yufan He}$^{2}$,
\textbf{Can Zhao}$^{2}$, \textbf{Daguang Xu}$^{2}$, \textbf{Yucheng Tang}$^{2}$, \textbf{Yuyin Zhou}$^{1}$\\
$^{1}$University of California, Santa Cruz \quad $^{2}$NVIDIA
}

\newif\ifarxiv
\arxivtrue
\ifarxiv\iclrfinalcopy\fi
\begin{document}

\maketitle
\ifarxiv
\lhead{Preprint}
{\let\thefootnote\relax\footnotetext{$^{*}$Equal technical contribution.}}
\fi

\begin{abstract}
Large language models (LLMs) and agentic systems have shown promise for clinical decision support, but existing works largely assume that evidence has already been curated and handed to the model. 
Real-world clinical workflows instead require agents to actively seek, iteratively plan, and synthesize multimodal evidence from heterogeneous sources. 
In this paper, we introduce \textbf{ClinSeekAgent}, an automated agentic framework for dynamic multimodal evidence seeking that shifts the paradigm from passive evidence consumption to active evidence acquisition. 
Given a clinical task and access to raw patient data, ClinSeekAgent queries longitudinal EHR records and invokes imaging analysis tools when relevant; refines its hypotheses as new information emerges; and integrates the collected evidence into grounded clinical decisions.
We validate ClinSeekAgent in two roles: as an inference-time agent for frontier LLMs and as a training pipeline for open-source models.
For inference-time validation, we construct \textbf{ClinSeek-Bench}, which covers 13 text-only EHR task categories and 6 multimodal EHR--chest X-ray task categories, and pairs each question with both a \textit{Curated Input} of pre-selected evidence and the raw patient data for automated evidence seeking.
Without any benchmark-specific curation, ClinSeekAgent significantly outperforms curated input for both Claude Opus 5 and GPT-5.6 Sol, improving overall F1 from 61.7 to 63.9 and from 57.8 to 59.3 on text-only EHR tasks, and from 66.5 to 68.8 and from 60.0 to 62.5 on multimodal tasks.
The largest gains occur in text-based decision-making and multimodal phenotype prediction, and ablations show that they come from multi-round evidence seeking rather than from access to more of the patient record.
For training-time validation, we fine-tune Qwen3.5-35B-A3B on ClinSeekAgent trajectories generated by GPT-5.6 Sol.
The resulting \textbf{ClinSeek-35B-A3B} reaches 35.6 average F1 on AgentEHR-Bench (\textbf{+16.1} over its base model), outperforming all evaluated open-source models, including ones more than ten times larger, as well as Claude Opus 5.
\ifarxiv
Our model, data, and code are available at \url{https://github.com/UCSC-VLAA/ClinSeekAgent}.
\else
We will release our model, data, and code to facilitate future research.
\fi
\end{abstract}

\input{sections/intro}
\input{sections/methodology}

\input{sections/benchmark}

\input{sections/training}
\input{sections/related_works}
\input{sections/conclusion}

\subsection*{AI use statement}

Generative AI plays two distinct roles in this work.
First, large language models are part of the systems and pipelines we study, as described in the paper: frontier LLMs act as the agent of \Method{} and as the model under \Baseline{} (\cref{sec:inference_validation}); GPT-5.6 Sol generates the \Method{} trajectories used as synthetic supervision for training ClinSeek-35B-A3B (\cref{sec:training_exp}); and Claude Sonnet 4.6 serves as an answer judge for single-label multimodal questions (\cref{app:metric}).
Second, the authors used an AI coding assistant (Claude, through Claude Code) for research support.
It was used to implement and debug experiment code, including the agent runners, tool servers, data-processing and scoring scripts; to write the scripts that compute the reported statistics (paired bootstrap confidence intervals, multi-run consistency, record-length stratification, and trajectory behavior analysis); and to draft and edit parts of the paper.
We have not used generative AI tools to propose the research idea or hypotheses of this work, to create benchmark labels, or to develop theoretical models or mathematical claims; translation, surveys, interviews, and transcription are not applicable to this work.
We reviewed all AI-assisted work.
Every number reported in the paper is computed by scripts from the raw run outputs and was cross-checked against those outputs by the authors; each case study was verified against the original trajectories; AI-written code was reviewed and tested by the authors; and AI-drafted text was revised by the authors, with explanatory claims restricted to those supported by our experiments.
We take responsibility for the final content of this work, including text, claims, or artifacts produced with the aid of generative AI.




\subsection*{Reproducibility statement}

\textbf{Data.}
All benchmarks are built from public, credentialed-access datasets (MIMIC-IV and MIMIC-CXR on PhysioNet) and existing benchmarks (EHR-Bench, EHRXQA, MedMod, and AgentEHR-Bench).
Because the patient data cannot be redistributed, we will release the scripts that rebuild \Benchmark{} from these sources, together with the question identifiers of every sampled example.
\Cref{sec:clinseek_bench_construction} describes how the benchmark is constructed, \cref{tab:taxonomy} lists its task categories, and \cref{app:temporal_policy} specifies the temporal policy that determines which records the model may see; \cref{app:source_temporal_leakage} compares this policy with the temporal settings of the source benchmarks.

\textbf{Inference and evaluation.}
The tool space of \Method{} is described in \cref{sec:tool_space} and listed in \cref{app:tool_schema}.
\Cref{app:eval_details} gives the metric and its computation, the inference settings of every model (\cref{tab:eval_hyperparams}), and how runs that end without an answer are handled.
Statistical significance is assessed with a paired bootstrap over questions using a fixed random seed.
Because of API cost, run-to-run consistency is verified on one model, GPT-5.6 Sol, with five independent runs per inference mode (\cref{app:multirun}).
Closed-source models are accessed through provider APIs, whose model versions may change over time; we will release the raw outputs of all reported runs so that every table and figure can be recomputed without re-querying these models.

\textbf{Training.}
\Cref{sec:training_settings} and \cref{app:sft_config} describe the training data and the full fine-tuning configuration of ClinSeek-35B-A3B, which is evaluated with the same inference settings as its base model.
We will release the model weights, the training trajectories, and the code for the agent, training, scoring, and analysis.



\bibliography{iclr2027_conference}
\bibliographystyle{iclr2027_conference}

\input{sections/appendix}
\end{document}

%% file: math_commands.tex
\usepackage{amsmath,amsfonts,bm}

\def\eqref#1{equation~\ref{#1}}

\def\1{\bm{1}}

\DeclareMathAlphabet{\mathsfit}{\encodingdefault}{\sfdefault}{m}{sl}
\SetMathAlphabet{\mathsfit}{bold}{\encodingdefault}{\sfdefault}{bx}{n}



%% file: sections/intro.tex
\section{Introduction}
\label{sec:intro}

Recent large language models (LLMs) and agentic systems have shown strong potential in medical question answering, diagnostic reasoning, and clinical decision support~\citep{wu2025medreason,kim2024mdagents,fallahpour2025medrax,yao2022react,schmidgall2024agentclinic,zhang2023huatuogpttaminglanguagemodel}. 
However, many existing medical-agent settings remain overly simplistic, deviating from real-world clinical workflows. 
They often rely on general medical knowledge~\cite{wu2025knowledge,zhao2025medrag} or short organized patient vignettes, whereas real-world clinical decision support requires actively seeking patient-specific longitudinal information from raw Electronic Health Record (EHR) tables~\cite{johnson2016mimic,johnson2023mimic} and visual clues from medical imaging~\cite{johnson2019mimic}.
Such a limitation is particularly salient for clinical decision support, where the key challenge is not only to reason over given evidence, but also to decide where to retrieve evidence from, what evidence to retrieve, and how different pieces of evidence can be integrated into a grounded decision.

\begin{figure}[t]
\centering
{\includegraphics[width=0.9\textwidth]{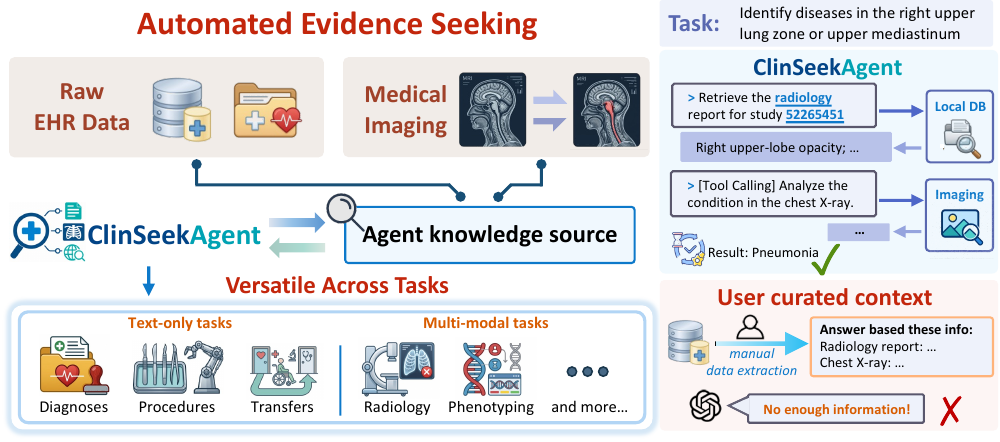}}
\captionof{figure}
{
\textbf{ClinSeekAgent Overview.}
ClinSeekAgent is an automated agentic evidence-seeking pipeline. It interacts with heterogeneous data sources to enable multimodal evidence seeking for clinical decision support. 
Compared with prior user-curated context settings, ClinSeekAgent is more flexible by acquiring richer information and knowledge from diverse tools.
}
\vspace{-8pt}
\vspace{-1.8em}
\label{fig:teaser}

\end{figure}

A growing line of EHR-specific work has moved closer to this goal by adapting LLMs to structured patient records and multimodal clinical data~\citep{liao2025ehr,bae2023ehrxqa,elsharief2025medmod,vasilev2025mtbbench}.
For example, recent EHR reasoning pipelines convert structured tables into textual contexts, retrieve task-related entities, and synthesize reasoning data from pre-extracted patient information~\citep{liao2025ehr,kweon2024ehrnoteqa}. 
Multimodal clinical benchmarks also combine EHRs and medical images to support realistic prediction and question-answering tasks~\citep{bae2023ehrxqa,elsharief2025medmod}. 
These efforts are valuable, but they still largely \emph{depend on a fixed evidence-packaging process} before inference: the relevant patient context is selected by benchmark construction, human priors, or task-specific rules. 
Recent studies of EHR agents have started to expose models to database tools~\citep{liao2026agentehr,medagentbench,medagentbenchv2,qian2026ehrnavigator,lee2025fhir,shi2024ehragent}, but they remain limited in task scope, tool coverage, or modality support. 
As a result, there is a need for a general agentic framework that automates the evidence search process, rather than assuming that the evidence has already been surfaced.

To address this need, we introduce \textbf{ClinSeekAgent}, an automated agentic framework for dynamic multimodal evidence seeking in clinical reasoning~(\cref{sec:method}). 
As shown in~\cref{fig:teaser}, ClinSeekAgent differs from existing curated-evidence pipelines in that it does not passively consume a fixed evidence package prepared before inference. 
Instead, given a clinical query and access to heterogeneous clinical data sources, ClinSeekAgent actively gathers evidence through \textbf{(1)} raw EHR retrieval and \textbf{(2)} medical imaging tools~(\cref{sec:tool_space}), iteratively refining its actions as new evidence emerges~(\cref{sec:evidence_seeking_trajectories}). 
This enables the agent to recover longitudinal and multimodal patient evidence that fixed curated contexts may miss.
For example, when asked to identify the diseases in the right upper lung zone or the upper mediastinum, \Method{} retrieves the radiology report of the study from the local EHR database, which notes a right upper-lobe opacity, and then analyzes the chest X-ray with an imaging tool to answer \textit{pneumonia}, while the same model under the curated-context setting fails due to missing critical evidence~(\cref{fig:teaser}).

We validate ClinSeekAgent first as an inference-time pipeline through \textbf{ClinSeek-Bench}, an evaluation suite that reformulates existing EHR and multimodal clinical tasks into paired curated-context and agentic settings~(\cref{sec:clinseek_bench_construction}). 
For each sample, the source benchmark~\cite{liao2025ehr,elsharief2025medmod,bae2023ehrxqa} provides a task-specific evidence package that was originally used as input to the model. 
We preserve this original setting as \textit{Curated Input}, where the model answers directly from the provided patient context. 
We then construct a paired \textit{Automated Evidence-Seeking} setting by removing this context and providing only the patient identifier, raw data access, and ClinSeekAgent tools, requiring the model to retrieve and integrate the necessary evidence by itself. 
As a result, each sample in ClinSeek-Bench evaluates the same task and answer label under two modes: answering from pre-selected evidence, and autonomously seeking evidence from raw clinical data. 
ClinSeek-Bench includes \textbf{13} text-only EHR task categories derived from EHR-Bench~\cite{liao2025ehr}, spanning risk prediction (mortality, readmission, length of stay, and escalation of care) and decision making (diagnosis, medication, procedure, laboratory and microbiology, vitals and measurements, imaging report, orders, encounter and flow, and next event), and \textbf{6} multimodal task categories adapted from EHRXQA~\citep{bae2023ehrxqa} and MedMod~\citep{elsharief2025medmod} (chest X-ray finding presence, finding enumeration, and change comparison, clinical deterioration, mortality, and phenotype)~(\cref{tab:taxonomy}).

Our inference-time experiments show that ClinSeekAgent can improve over fixed curated inputs when paired with capable agentic models. 
On text-only EHR tasks, overall F1 rises from 61.7 to 63.9 for Claude Opus 5 and from 57.8 to 59.3 for GPT-5.6 Sol when moving from \textit{Curated Input} to \textit{ClinSeekAgent}~(\cref{sec:exp_settings}, \cref{fig:category_gains}).
The largest gains occur in text-based decision-making tasks and multimodal phenotype prediction, where the agent must piece together evidence from longitudinal records and chest radiographs.
On the multimodal benchmark, these models improve from 66.5 to 68.8 and from 60.0 to 62.5 overall F1, respectively~(\cref{fig:category_gains}), suggesting that active evidence acquisition can recover clinical signals that fixed curated contexts may miss.

\begin{wrapfigure}{r}{0.60\textwidth}
    \vspace{-12pt}
    \centering
    \includegraphics[width=0.59\textwidth]{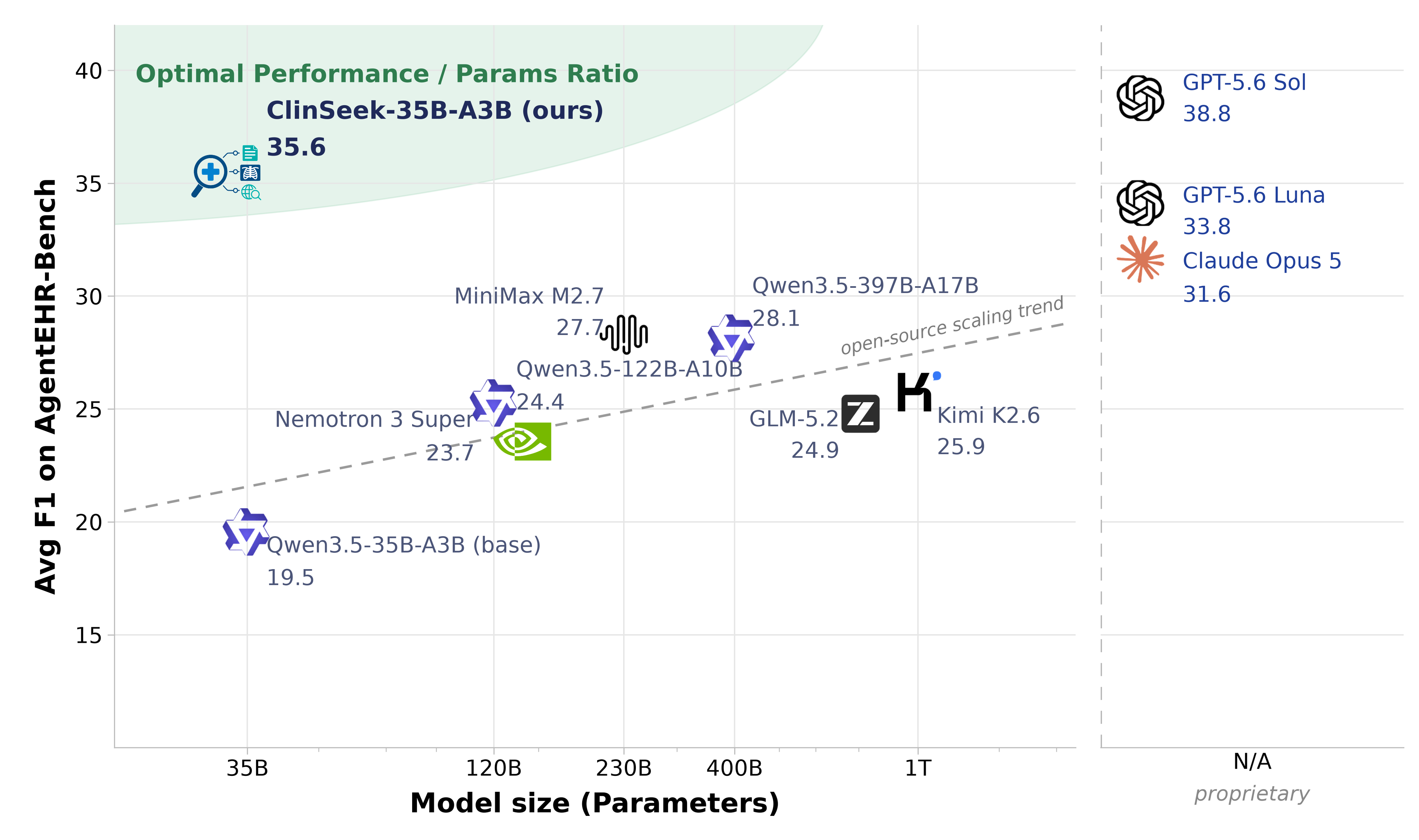}
    \vspace{-8pt}
    \caption{
    \textbf{Performance--model size comparison on AgentEHR-Bench.}
    ClinSeek-35B-A3B outperforms all open-source models, as well as Claude Opus 5.
    }
    \label{fig:agentehr_performance}
    \vspace{-10pt}
\end{wrapfigure}
Ablations show that iterative, multi-round evidence seeking drives the gains~(\cref{sec:gain_source}).
The variation across host models further suggests that this process depends on the agent's ability to plan and execute long-horizon tool use~(\cref{app:more_models}).
Therefore, we further validate ClinSeekAgent as a training pipeline for open-source clinical agents~(\cref{sec:training_exp}). 
Using ClinSeekAgent, we collect high-quality clinical search trajectories from a strong teacher model and fine-tune Qwen3.5-35B-A3B~\citep{qwen3.5}, resulting in \textbf{ClinSeek-35B-A3B}. 
On the existing AgentEHR-Bench~\cite{liao2026agentehr}, ClinSeek-35B-A3B improves over its base model from 19.5 to 35.6 average F1, outperforms all evaluated open-source models including Qwen3.5-397B-A17B (28.1), and exceeds Claude Opus 5 (31.6); among proprietary models only GPT-5.6 Sol scores higher (38.8)~(\cref{fig:agentehr_performance}, \cref{tab:SFT_performance}). 
These results show that ClinSeekAgent is not only effective as an inference-time pipeline, but can also serve as a scalable training pipeline for distilling clinical evidence-seeking behavior into open-source models~(\cref{sec:sft_behavior}).

%% file: sections/methodology.tex
\section{ClinSeekAgent: Multimodal Evidence-Seeking Pipeline}
\label{sec:method}


\subsection{Task Formulation and Interaction Protocol}
\label{sec:task_formulation}

Each clinical task instance is defined as:
\begin{equation}
    x = (p, t, q, \mathcal{M}, \mathcal{Y}),
\end{equation}
where $p$ is the patient identifier, $t$ is the reference timestamp or prediction time, $q$ is the clinical task instruction, $\mathcal{M}$ denotes optional modality-specific metadata such as image paths, and $\mathcal{Y}$ denotes the answer schema or candidate label space when available.
During inference, the model is not given the curated patient context used by the source benchmark.
Instead, it receives $x$ and access to the \Method tool space, and 
invokes tools to retrieve evidence needed for the task.
At step $k$, the model $\pi_\theta$ observes the task instance and the previous interaction history
\begin{equation}
    h_{k-1} = \{(a_1,o_1), \ldots, (a_{k-1},o_{k-1})\},
\end{equation}
and either invokes another tool or terminates the answering process as its next action:
\begin{equation}
    a_k \sim \pi_\theta(\cdot \mid x, h_{k-1}).
\end{equation}
If $a_k$ is a tool call, the environment returns an observation $o_k$; otherwise, the model outputs the final prediction $\hat{y}$ following the specified answer schema.
For EHR-related tasks, the agent first loads the patient database with \texttt{ehr.load\_ehr}, and all EHR queries are restricted to records available \textit{before the reference timestamp $t$}~(\cref{app:temporal_policy,app:source_temporal_leakage}).

\subsection{Multi-Source Tool Space}
\label{sec:tool_space}

\Method exposes a unified tool space with 22 tools across two complementary evidence sources: \textit{EHR retrieval} and \textit{medical image analysis}. 
Specifically, it provides 16 EHR tools for accessing patient-specific longitudinal records, including schema inspection, temporal, keyword-, and value-based retrieval, SQL-based querying, and candidate-term grounding; and 6 image tools for extracting visual evidence through DICOM preprocessing, image visualization, chest X-ray classification, report generation, phrase grounding, and anatomical segmentation. 
The complete tool list is provided in Appendix~\ref{app:tool_schema}.

\subsection{Agentic Evidence-Seeking Trajectories}
\label{sec:evidence_seeking_trajectories}

\Method represents each run as an open-ended evidence-seeking trajectory:
\[
\tau = \bigl(x,\,(a_k,o_k)_{k=1}^{K},\,\hat{y}\bigr)
\]
where $x$ is the task instance, $a_k$ is a tool action, $o_k$ is the corresponding tool observation, and $\hat{y}$ is the final answer.
The trajectory records both the final prediction and the sequence of evidence-seeking decisions that produced it.

Unlike rule-based retrieval pipelines, \Method does not impose an ordering over evidence sources.
Depending on the task, the model may begin with schema inspection, EHR querying, image analysis, or candidate retrieval, and may interleave these tools across multiple turns.
Thus, \Method standardizes the environment and tool interface, while the evidence-seeking policy is induced by the agentic model.

%% file: sections/benchmark.tex
\section{Inference-time Validation: Curated Input vs Automated Evidence Seeking}
\label{sec:inference_validation}

\subsection{ClinSeek-Bench Construction}
\label{sec:clinseek_bench_construction}

We construct ClinSeek-Bench to validate \Method{} as an inference-time evidence-seeking pipeline. 
Each example is paired into two settings with the same task definition and answer label: \textit{Curated Input}, where the model answers from the evidence package provided by the source EHR reasoning benchmark~\cite{liao2025ehr,bae2023ehrxqa, elsharief2025medmod}, and \textit{Automated Evidence-Seeking}, where this context is removed and the model must retrieve evidence from raw clinical data using \Method{} tools.

\paragraph{Source Benchmarks.}
\Benchmark{} has two parts.
\Benchmark-Text is built from EHR-Bench~\citep{liao2025ehr}, which contains 45 EHR analysis subtasks over MIMIC-IV; we randomly sample 40 examples from each subtask, resulting in 1,800 examples.
We group the subtasks into 13 task categories: 4 for risk prediction (mortality, readmission, length of stay, and escalation of care) and 9 for decision making (diagnosis, medication, procedure, laboratory and microbiology, vitals and measurements, imaging report, orders, encounter and flow, and next event).
\Benchmark-MM is built from EHRXQA~\citep{bae2023ehrxqa} and MedMod~\citep{elsharief2025medmod}, both of which pair MIMIC-IV EHRs with MIMIC-CXR chest radiographs. After reconstructing the official examples and preserving their task definitions, splits, labels, and EHR-CXR pairing rules, we obtain 989 examples across 6 task categories: CXR finding presence, CXR finding enumeration, CXR change comparison, clinical deterioration, mortality, and phenotype.
\Cref{tab:taxonomy} describes each task category.

\paragraph{Curated Input Data Collection.}
We preserve the \textit{original EHR reasoning benchmark inputs} as the \textit{Curated Input} setting. 
These inputs reflect the evidence-packaging process of the source benchmarks, where task-relevant patient information is selected before inference. 
For EHR-Bench, the original setting uses rule-based templates to convert recent patient events into instruction-answer samples: models observe up to 100 events from the past 24 hours and predict either the next clinical event or a future risk outcome. 
For EHRXQA and MedMod, we keep the original task-specific EHR context, selected CXR studies, image identifiers, labels, and pairing rules from the official repositories. 

\paragraph{Automated Evidence-Seeking Data Generation.}
We convert each curated example into an \textit{Automated Evidence-Seeking} example by removing the curated context while keeping the same task instruction and answer label. 
The model is instead given the patient identifier, prediction-time cutoff, optional linked CXR identifiers, and access to ClinSeekAgent tools. 
For EHR-Bench, we use the timestamp of the last event in the original input as the reference cutoff, allowing the agent to access the patient's full raw EHR history before that time. 
For multimodal tasks, we preserve the original patient-level task, label, and valid EHR-CXR linkage, but require the agent to retrieve EHR evidence and invoke imaging tools when needed. 

\paragraph{Temporal Filtering.}
Both data types are filtered by the same strict temporal policy: the model can only see records that were available before the prediction cutoff.
Because diagnosis and procedure codes are assigned when an encounter is completed, they are treated as unavailable until then, and the codes of the encounter being predicted are always hidden. More details are provided in \cref{app:temporal_policy}, and \cref{app:source_temporal_leakage} compares this policy with the temporal settings of the source benchmarks.

\subsection{Main Results: \Method~Improves Frontier Agentic Models}
\label{sec:exp_settings}

\begin{figure}[t]
\centering
\includegraphics[width=\linewidth]{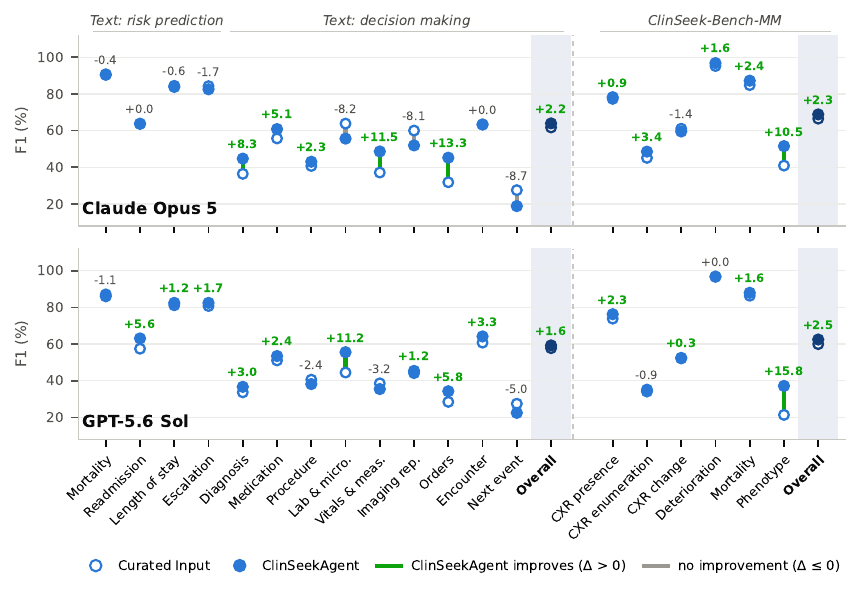}
\vspace{-2.2em}
\caption{
\textbf{\Method{} vs.\ \Baseline{} by task category} (sample-wise F1, \%) for Claude Opus 5 (top) and GPT-5.6 Sol (bottom) on \Benchmark-Text (left) and \Benchmark-MM (right).
Hollow markers are the curated input, filled markers are \Method, and the label above each pair gives their difference; \textcolor[HTML]{0ca30c}{green} marks categories where \Method{} improves.
Categories follow the task taxonomy in \cref{app:more_models}; the shaded column is the overall score.
}
\vspace{-.5em}
\label{fig:category_gains}
\end{figure}

\paragraph{Experimental Settings.} We compare two inference modes on the paired data of \cref{sec:clinseek_bench_construction}.
In the \Baseline{} mode, the pre-prepared evidence package is rendered in the prompt and the model answers from it.
\Method{} runs in the Automated Evidence-Seeking setting: given only the patient identifier and the prediction cutoff, the model retrieves evidence from the raw record with the tools of \cref{sec:tool_space} for up to 200 rounds before calling \texttt{finish}.
Both modes run with the same reasoning budget and one run per question; the remaining settings are listed in \cref{app:eval_details}.
Since most tasks are multi-label prediction, we use sample-wise F1 (\%) as the unified metric across all tasks; its computation is detailed in \cref{app:metric}.
We evaluate two frontier models, Claude Opus 5~\cite{Opus5} and GPT-5.6 Sol~\cite{GPT5.6}, and report both modes and their difference per task category in \cref{fig:category_gains}; results for Claude Opus 4.6~\cite{Opus4.6}, GLM-5.2~\cite{zeng2025glm}, and Kimi K2.6~\cite{team2026kimi} are given in \cref{app:more_models}.

\paragraph{Automatic evidence seeking replaces hand-designed curation.}
As shown in \cref{fig:category_gains}, switching from \Baseline{} to \Method{} raises the overall F1 of Claude Opus 5 from 61.7 to 63.9 on \Benchmark-Text and from 66.5 to 68.8 on \Benchmark-MM, and that of GPT-5.6 Sol from 57.8 to 59.3 and from 60.0 to 62.5, respectively.
The gains are broad rather than driven by a single task: \Method{} improves 23 of the 38 model--category pairs, including 13 of 19 categories for GPT-5.6 Sol and 10 of 19 for Claude Opus 5, with the largest gains reaching +13.3 (orders) and +11.5 (vitals and measurements) for Opus 5 and +15.8 (phenotype) and +11.2 (laboratory and microbiology) for Sol; the largest decreases ($-8$ to $-9$ points for Opus 5) fall in the three smallest categories, with 40--80 questions each.

\Method{} achieves this \textit{without the benchmark-specific curation rules} that \Baseline{} relies on: the agent decides what to retrieve from the raw record.
Models with stronger agentic capability benefit more from the pipeline: Claude Opus 4.6, GLM-5.2, and Kimi K2.6 improve by smaller margins, +5.1 (Opus 4.6) and +2.0 (Kimi K2.6) on \Benchmark-MM and +0.7 (GLM-5.2) on \Benchmark-Text, and only the gain of Opus 4.6 is significant (\cref{app:more_models}).

\paragraph{Reasoning-intensive tasks benefit the most.}
The largest improvements appear on the decision-making tasks of \Benchmark-Text and on phenotype prediction in \Benchmark-MM.
Decision-making tasks ask which clinical events come next, e.g., the diagnoses, medications, or orders, so the model must relate the findings in the record to the patient's course and enumerate a set of concrete items.
Predicting the next diagnoses, medications, and orders improves for both models (+8.3/+5.1/+13.3 for Opus 5, +3.0/+2.4/+5.8 for Sol); in addition, Opus 5 gains +11.5 on vitals and measurements and Sol gains +11.2 on laboratory and microbiology.
Phenotype prediction asks for the set of chronic and acute conditions of an ICU stay, which has to be assembled from the radiograph together with several EHR tables; it improves by +10.5 for Opus 5 and +15.8 for Sol.
Risk prediction, in contrast, only asks for a binary label and is already close to saturation under \Baseline{}; \Method{} maintains this high level on both benchmarks (\Method{} vs.\ \Baseline{}: 81.7 vs.\ 82.2 for Opus 5 and 79.6 vs.\ 78.2 for Sol over the 720 risk-prediction questions of \Benchmark-Text, and 87--97 F1 on decompensation and mortality in \Benchmark-MM).

\paragraph{The improvements are stable across multiple runs.}
To check that the differences above are not run-to-run noise, we repeat both inference modes for GPT-5.6 Sol four more times on both benchmarks (\cref{app:multirun}).
Across the five runs, the overall F1 has a standard deviation of at most 0.4 points within each mode (\Method{} vs.\ \Baseline{}: 59.4$\pm$0.1 vs.\ 57.4$\pm$0.2 on \Benchmark-Text; 62.4$\pm$0.4 vs.\ 59.7$\pm$0.3 on \Benchmark-MM), and \Method{} beats \Baseline{} in every run, by +1.6 to +2.3 and +2.3 to +3.3 points.
At the question level, the runs agree closely: Krippendorff's $\alpha$ of per-question F1 across runs is 0.87--0.91 for both modes (metric definitions in \cref{app:consistency_metrics}).

\subsection{Where Does the Gain Come From?}
\label{sec:gain_source}
\Method{} differs from \Baseline{} in two ways: (1) the model can access the whole record instead of a curated window; and (2) it gathers evidence from the record over multiple rounds of tool calls and reasoning rather than in a single step.
To ablate the two factors, we add two settings that provide extended record access in different ways but without the multi-round loop (\cref{tab:four_modes}).
\textit{Full-Record Input} is a variant of curated input: the curated evidence package is replaced by the patient's full pre-cutoff record, rendered into the prompt in the same way and truncated oldest-first when it does not fit in the 100k-token context window.
\textit{Single-Pass Retrieval} is a variant of \Method{}: the agent keeps the same tools and access to the whole record, but must issue all of its retrieval calls in one turn and answer from what comes back.
All four settings share the temporal policy of \cref{sec:clinseek_bench_construction}~(\cref{app:temporal_policy,app:source_temporal_leakage}).

\input{tables/table_four_modes}

\begin{figure}[t]
\centering
\vspace{-0.5em}
\includegraphics[width=\linewidth]{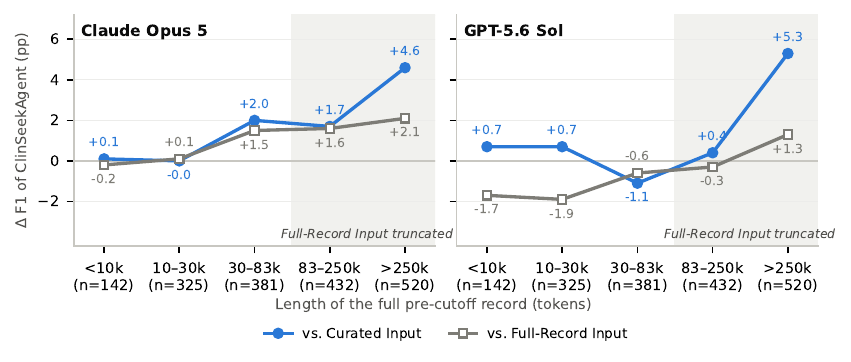}
\vspace{-2.5em}
\caption{\textbf{Gain of \Method{} grows with the length of the patient record} (\Benchmark-Text, F1 points). Questions are binned by the token count of the full pre-cutoff record; in the shaded bins the record does not fit in the 100k-token context window of Full-Record Input and is truncated.
}
\label{fig:length_strata}
\end{figure}

\paragraph{Extended record access alone does not help.}
Neither way of extending record access recovers the gain of \Method{}.
Full-Record Input gives the model the whole record instead of the curated window, yet its overall F1 exceeds that of \Baseline{} by only +0.8/+1.9 points on \Benchmark-Text and +0.1/+0.3 on \Benchmark-MM (Opus 5/Sol).
The length of the record makes this concrete.
\Cref{fig:length_strata} stratifies the \Benchmark-Text questions by the length of the patient's full pre-cutoff record: below 30k tokens \Method{} is on par with both prompt-based settings, and its margin grows with record length to +4.6/+5.3 over \Baseline{} and +2.1/+1.3 over Full-Record Input on records above 250k tokens, where the prompt can no longer hold the record and a fixed curation rule has the most to miss.
A case study (\cref{fig:case_full_record} in \cref{app:case_full_record}) shows that having the record in the prompt is not the same as finding the deciding row: the unit to which the same patient was admitted after a comparable orthopedic visit two weeks earlier sits at 26\% of the untruncated Full-Record prompt, yet the model answers with a string near the end of the prompt that is not a care unit in the record.
\Method{} instead notices that the admission transfer is still pending, widens its time window, and retrieves the precedent.

Single-Pass Retrieval can access the entire record, but it is no better than \Baseline{} either: its overall F1 differs from that of \Baseline{} by --3.2/--0.1 on \Benchmark-Text and --3.9/+1.2 on \Benchmark-MM, with no significant gain.
Reading a longer prompt or issuing all queries in one turn thus adds little over \Baseline{}.

\begin{figure}[t]
\centering
\vspace{-1em}
\includegraphics[width=\linewidth]{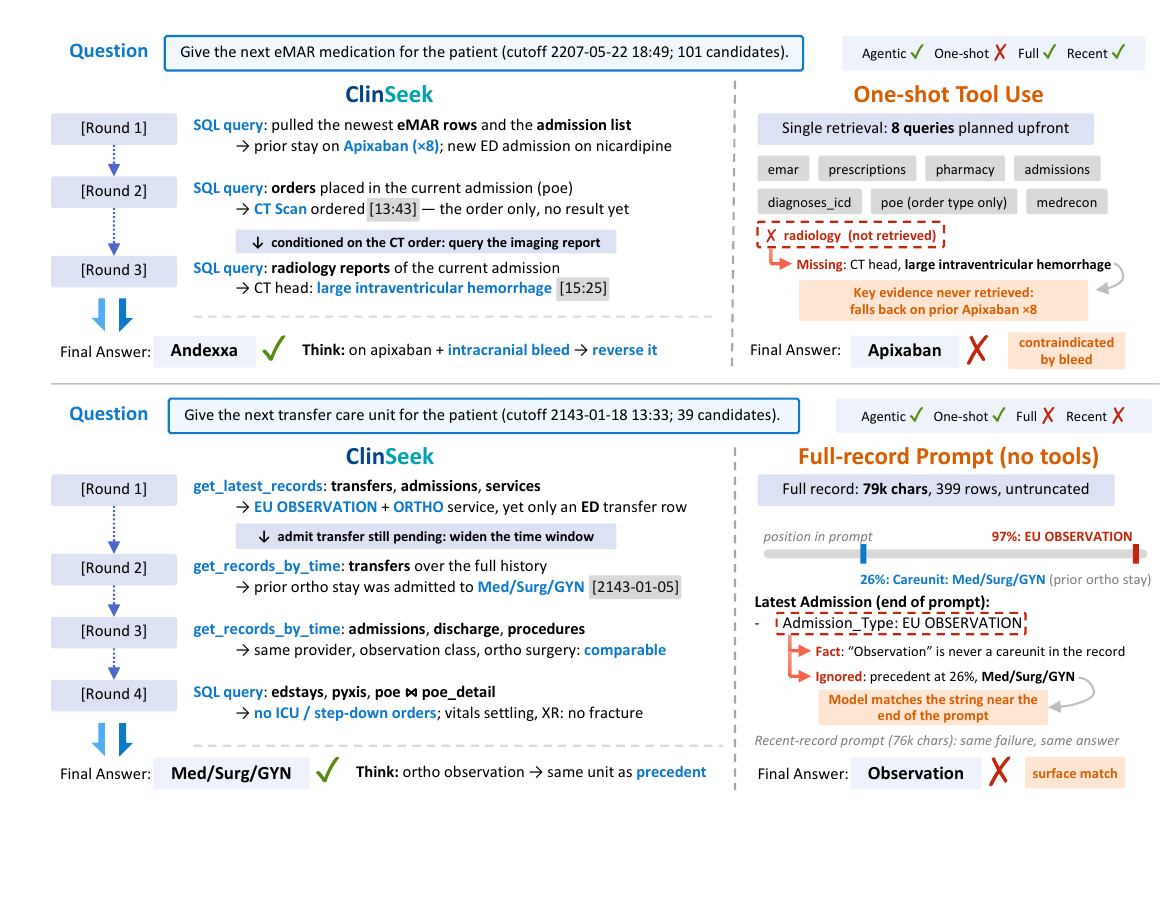}
\vspace{-1.5em}
\caption{\textbf{Case study: \Method{} vs.\ Single-Pass Retrieval}.
}
\vspace{-1.7em}
\label{fig:case_single_pass}
\end{figure}

\paragraph{Multi-round reasoning is the source of the gain.}
Single-Pass Retrieval and \Method{} differ only in whether retrieval is iterative, so their gap measures the effect of multi-round reasoning alone: +5.4/+1.7 on \Benchmark-Text and +6.1/+1.4 on \Benchmark-MM.
Each round lets the agent condition its next query on what the previous one returned, which neither variant can do.
\Cref{fig:case_single_pass} shows this on one question: the first round returns the patient's prior anticoagulant and a pending CT order, so the agent requests the CT report, finds a large intracranial bleed, and predicts the reversal agent.
Single-Pass Retrieval had to plan all of its queries before seeing any result and thus without that pointer; it never retrieved the report and predicted the anticoagulant that the bleed contraindicates.

%% file: tables/table_four_modes.tex
\providecommand{\gapneg}[1]{\textcolor{red!60!black}{--#1}}
\providecommand{\gappos}[1]{\textcolor{green!40!black}{+#1}}
\begin{table}[t]
\centering
\small
\setlength{\tabcolsep}{5pt}
\caption{\textbf{Ablating the two ingredients of \Method{}} (overall F1, \%).
Rounds: model calls per question. In parentheses: that setting -- \Method{}, \textcolor{red!60!black}{red} for a decrease and \textcolor{green!40!black}{green} otherwise, bold when the paired-bootstrap 95\% CI excludes zero.}
\label{tab:four_modes}
\vspace{-5pt}
\begin{tabular}{lccc|cc|cc}
\toprule
& & & & \multicolumn{2}{c|}{\Benchmark-Text} & \multicolumn{2}{c}{\Benchmark-MM} \\
\textbf{Setting} & Record & Tools & Rounds & Opus 5 & Sol & Opus 5 & Sol \\
\midrule
\Baseline & recent & -- & 1 & 61.7 (\textbf{\gapneg{2.2}}) & 57.8 (\textbf{\gapneg{1.6}}) & 66.5 (\textbf{\gapneg{2.3}}) & 60.0 (\textbf{\gapneg{2.5}}) \\
Full-Record Input & full & -- & 1 & 62.6 (\gapneg{1.3}) & 59.6 (\gappos{0.3}) & 66.6 (\textbf{\gapneg{2.2}}) & 60.3 (\textbf{\gapneg{2.3}}) \\
Single-Pass Retrieval & full & \checkmark & 2 & 58.5 (\textbf{\gapneg{5.4}}) & 57.6 (\textbf{\gapneg{1.7}}) & 62.6 (\textbf{\gapneg{6.1}}) & 61.2 (\gapneg{1.4}) \\
\Method & full & \checkmark & 5--10 & 63.9 & 59.3 & 68.8 & 62.5 \\
\bottomrule
\end{tabular}
\vspace{-15pt}
\end{table}

%% file: sections/training.tex
\section{Training-time Validation: Teaching Open Models to Use ClinSeekAgent}
\label{sec:training_exp}
\Cref{sec:gain_source} shows that the gain of \Method{} comes from multi-round evidence seeking, and \cref{sec:exp_settings} shows that this gain is consistent only for models with strong agentic capability.
We therefore ask whether \Method{} can also serve as a training pipeline that transfers this capability to a smaller open-source model.
Specifically, we use the trajectories that \Method{} produces with a strong teacher as supervision, and test whether the student learns the evidence-seeking process rather than only the final answers.

\subsection{Experimental Settings}
\label{sec:training_settings}
We use GPT-5.6 Sol as the teacher to run \Method{} on the 3,000 questions of the AgentEHR-Bench~\cite{liao2026agentehr} training split, and fine-tune Qwen3.5-35B-A3B on the resulting trajectories with supervised fine-tuning.
The teacher's intermediate reasoning is placed in the student's native thinking block, and the tool calls are rendered in its native tool-call format (\cref{app:sft_config}).
We evaluate on its test split, using 100 questions for each of its five tasks; none of these questions appears in the training data.
All models run \Method{} with the same tools, temporal policy~(\cref{app:temporal_policy,app:source_temporal_leakage}), and inference settings as in \cref{sec:exp_settings}; the base and fine-tuned models are served locally with identical decoding settings.

\subsection{ClinSeek-35B-A3B Achieves Open-Source State-of-the-Art}

\Cref{tab:SFT_performance} reports the results on AgentEHR-Bench.
Training on \Method{} trajectories raises the average F1 of Qwen3.5-35B-A3B from 19.5 to 35.6 (\textbf{+16.1}).
The model improves on all five tasks, by +7.6 on \textit{Microbiology} up to +23.9 on \textit{Labs} and +23.4 on \textit{Procedures}, and every gain is significant under a paired bootstrap.
ClinSeek-35B-A3B outperforms every open-source model we evaluate, including Qwen3.5-397B-A17B, which has about eleven times more parameters, by +7.5 points.
It also exceeds Claude Opus 4.6 by +2.4 points and Claude Opus 5 by +4.0, and reaches 92\% of its teacher GPT-5.6 Sol (35.6 vs.\ 38.8), the only closed-source model that scores higher; on \textit{Transfers} it matches the teacher (27.4 vs.\ 27.3).
These results show that \Method{} trajectories transfer long-horizon EHR evidence seeking to a smaller open-source model.

\input{tables/SFT_performance}

\subsection{What Does ClinSeek-35B-A3B Learn?}
\label{sec:sft_behavior}
To understand where the gain comes from, we compare the 500 trajectories of ClinSeek-35B-A3B with those of its base model (\cref{app:sft_behavior}).
Both precision and recall improve (18.6 to 32.4 and 31.1 to 53.3).
Two changes in behavior accompany this gain:
\begin{itemize}
\item \textbf{More complex SQL queries.} The student learns to write SQL with joins, aggregation, or time filters, which it uses in 53\% of the questions compared with 0.8\% for the base model; it typically first summarizes what the record contains and then drills into the current admission.
\item \textbf{Reasoning grounded on the patient record.} The student surveys the record before narrowing down: it reads twice as many of the available tables (72\% vs.\ 36\%), issues a larger share of targeted follow-up queries (36\% vs.\ 13\%), reads imaging or discharge notes in 74\% vs.\ 12\% of the questions, and refers to the current admission in its reasoning in 80\% vs.\ 39\%.
\end{itemize}
The student thus learns the evidence-seeking process using \Method{}, not only its final answers.

%% file: tables/SFT_performance.tex
\begin{table*}[t]
\centering
\small
\setlength{\tabcolsep}{5pt}
\caption{\textbf{Results on AgentEHR-Bench} (F1, \%; 100 questions per task). All models run \Method{} under the same temporal policy. Bold marks the best model in each group. Last two rows: difference of ClinSeek-35B-A3B from its base model and from its teacher, \textcolor{red!60!black}{red} for a decrease and \textcolor{green!40!black}{green} otherwise; the gains over the base model are all significant under a paired bootstrap (95\% CI).}
\label{tab:SFT_performance}
\begin{tabular}{lcccccc}
\toprule
\textbf{Model} & \textbf{Diagnoses} & \textbf{Labs} & \textbf{Microbiology} & \textbf{Procedures} & \textbf{Transfers} & \textbf{Avg.} \\
\midrule
\multicolumn{7}{l}{\textit{Closed-source models}} \\
GPT-5.6 Sol (teacher) & 47.0 & 43.1 & \textbf{31.1} & \textbf{45.5} & \textbf{27.3} & \textbf{38.8} \\
GPT-5.6 Luna & 43.7 & 37.8 & 24.8 & 37.9 & 24.9 & 33.8 \\
Claude Opus 4.6 & \textbf{51.1} & 42.1 & 25.4 & 28.7 & 18.6 & 33.2 \\
Claude Sonnet 4.6 & 49.5 & 36.8 & 22.9 & 25.9 & 24.5 & 31.9 \\
Claude Opus 5 & 48.0 & \textbf{44.5} & 23.2 & 20.5 & 22.0 & 31.6 \\
\midrule
\multicolumn{7}{l}{\textit{Open-source models}} \\
Qwen3.5-397B-A17B & 38.4 & 30.9 & 21.2 & 26.2 & 23.7 & 28.1 \\
MiniMax M2.7 & 39.4 & 22.3 & 22.1 & 32.0 & 22.9 & 27.7 \\
Kimi K2.6 & 42.8 & 20.2 & 18.5 & 24.6 & 23.3 & 25.9 \\
GLM-5.2 & \textbf{43.8} & 20.4 & 16.0 & 24.0 & 20.4 & 24.9 \\
Qwen3.5-122B-A10B & 32.6 & 23.7 & 21.6 & 23.8 & 20.5 & 24.4 \\
Nemotron 3 Super & 29.5 & 27.5 & 23.1 & 24.9 & 13.3 & 23.7 \\
Qwen3.5-35B-A3B (base) & 28.4 & 16.5 & 19.5 & 17.9 & 15.1 & 19.5 \\
\textbf{ClinSeek-35B-A3B (ours)} & 41.7 & \textbf{40.4} & \textbf{27.1} & \textbf{41.3} & \textbf{27.4} & \textbf{35.6} \\
\midrule
Ours -- base & \textcolor{green!40!black}{+13.3} & \textcolor{green!40!black}{+23.9} & \textcolor{green!40!black}{+7.6} & \textcolor{green!40!black}{+23.4} & \textcolor{green!40!black}{+12.3} & \textcolor{green!40!black}{+16.1} \\
Ours -- teacher & \textcolor{red!60!black}{--5.3} & \textcolor{red!60!black}{--2.7} & \textcolor{red!60!black}{--4.0} & \textcolor{red!60!black}{--4.2} & \textcolor{green!40!black}{+0.1} & \textcolor{red!60!black}{--3.2} \\
\bottomrule
\end{tabular}
\vspace{-2em}
\end{table*}

%% file: sections/related_works.tex
\section{Related Work}

\paragraph{Medical Reasoning with Curated Evidence.}
Recent medical LLMs have shown strong performance in medical question answering and diagnostic reasoning~\cite{tu2024towards,ossowski2025octomed,huang2025medvlthinker,huang2025m1,li2023llava,shi2026medxiaohe,wang2026deepmed}, demonstrating that LLMs can encode medical knowledge and reason over clinical scenarios. 
These settings differ from real-world clinical decision support, where models must first identify and retrieve task-relevant evidence from longitudinal patient records, rather than only reason over provided patient vignettes~\cite{jin2019pubmedqa,jin2021disease}, summarized clinical notes~\cite{kweon2024ehrnoteqa}, or task-specific patient contexts~\cite{yu2025medframeqa,zuo2025medxpertqa}. 
Recent EHR and multimodal clinical benchmarks move closer to real clinical data by grounding tasks in structured patient records, radiology reports, and medical images~\cite{liao2025ehr,elsharief2025medmod,bae2023ehrxqa}. 
However, these works still largely follow the curated-evidence paradigm: task-relevant records, reports, or multimodal inputs are selected before inference. 
In contrast, \Method{} focuses on automating this evidence-seeking step, allowing the agent to dynamically query raw EHR tables and invoke medical image analysis tools.

\paragraph{Agentic Evidence Seeking over Clinical Data.} Recent medical agent systems have begun to move beyond single-pass reasoning by introducing tool use, search, and multi-agent collaboration into clinical tasks. 
MDAgents adaptively organizes multiple LLM agents for medical decision making~\cite{kim2024mdagents}, while DeepMed~\cite{wang2026deepmed} and Meissa~\cite{chen2026meissa} train medical agents to perform multi-step evidence search or interaction for medical reasoning.
Closer to EHR-based decision support, AgentEHR~\cite{liao2026agentehr}, MedAgentBench~\cite{medagentbench}, and FHIR-AgentBench~\cite{lee2025fhir} evaluate agents in interactive clinical record environments, requiring models to retrieve patient information and reason over structured records. 
AgentClinic further studies tool-using agents in simulated multimodal clinical environments~\cite{schmidgall2024agentclinic}. 
These works demonstrate the promise of agentic clinical AI, but their evidence-seeking processes are typically limited to medical knowledge search, multi-agent discussion, EHR-only interaction, or simulated clinical tools. 
In contrast, \Method{} provides a unified multimodal evidence-seeking pipeline over raw EHR tables and medical image analysis tools, and further validates this pipeline both at inference time and through trajectory-based training of open-source agents.

%% file: sections/conclusion.tex
\section{Conclusion}

In this paper, we introduce \Method, an automated agentic framework for dynamic multimodal evidence seeking in clinical decision support, which lets an agentic model gather evidence from raw EHR records and chest X-ray analysis tools instead of relying on curated inputs.
As an inference-time pipeline, \Method{} improves two frontier models over curated inputs on both text-only and multimodal clinical tasks without any benchmark-specific curation, with the largest gains on reasoning-intensive tasks.
Ablations show that these gains come from multi-round evidence seeking rather than from access to more of the patient record, and that they grow with the length of the record.
As a training pipeline, \Method{} trajectories teach an open-source model, ClinSeek-35B-A3B, to write more complex queries and to ground its reasoning in the patient record, outperforming all evaluated open-source models on AgentEHR-Bench.
Our results suggest that moving from passive evidence consumption to active evidence acquisition is a promising direction for building more flexible, grounded, and capable clinical AI agents.

%% file: sections/appendix.tex
\appendix
\section*{Technical Appendix}
\section{Limitations and Discussion}
\label{sec:limitation}
While \Method{} demonstrates promising results as both an inference-time and training-time pipeline, several limitations remain. 
First, the current multimodal evaluation tasks are still relatively simple in many cases. 
Although they involve both EHR and imaging evidence, many examples can be solved with a small number of tool calls or with limited cross-modal interaction. 
This does not fully stress-test the long-horizon multimodal evidence-seeking capability that \Method{} is designed for. 
Future benchmarks should include more challenging clinical scenarios where the agent must iteratively combine raw EHR retrieval, medical image analysis, external knowledge, and temporal reasoning over extended patient histories.

Second, our current training pipeline relies primarily on supervised fine-tuning over teacher-generated trajectories. 
Teacher trajectories may contain redundant or low-value tool calls, which can pollute the context window and teach the student suboptimal evidence-seeking behavior. 
Improving the quality of teacher trajectories through refinement, filtering, or compression is therefore an important direction for future work. 
In addition, post-SFT reinforcement learning could further improve the model's generalization, efficiency, and robustness by directly optimizing successful and concise clinical evidence seeking rather than merely imitating teacher behavior. 
We are actively working on these directions to build more challenging evaluations and more efficient training pipelines for clinical evidence-seeking agents.

\section{Results for More Models}
\label{app:more_models}
\Cref{tab:text_appendix,tab:mm_appendix} report \Method{} and \Baseline{} for Claude Opus 4.6, GLM-5.2, and Kimi K2.6 under the same protocol as \cref{sec:exp_settings}.
On \Benchmark-Text all three stay within two points of \Baseline{} overall; on \Benchmark-MM both evaluated models improve, Claude Opus 4.6 mostly on phenotype prediction and Kimi K2.6 mostly on CXR finding presence and change comparison.
\Cref{tab:taxonomy} lists the task categories used in \cref{fig:category_gains}.
\input{tables/table_text_appendix}
\input{tables/table_mm_appendix}

\begin{table}[t]
\centering
\small
\caption{\textbf{Task categories of \Benchmark.} Each \Benchmark-Text task is mapped to one category; \Benchmark-MM categories follow the source task groups.}
\label{tab:taxonomy}
\begin{tabularx}{\linewidth}{@{}l >{\raggedright\arraybackslash}X@{}}
\toprule
\textbf{Category} & \textbf{Tasks} \\
\midrule
\multicolumn{2}{l}{\textit{\Benchmark-Text, risk prediction}} \\
Mortality & Predict whether the patient dies during hospitalization (from an ED visit or an inpatient stay), or within 1, 2, 3, 7, or 14 days in the ICU \\
Readmission & Predict whether the patient returns to the ED within 72 hours, is readmitted to the ICU during the same hospitalization, or is readmitted to the hospital within 30 or 60 days \\
Length of stay & Predict whether the ICU stay exceeds 7 or 14 days, or the hospital stay exceeds 3 or 7 days \\
Escalation of care & Predict whether an ED patient is hospitalized, is transferred to the ICU within 12 hours, or has a critical outcome (in-hospital death or ICU transfer within 12 hours) \\
\multicolumn{2}{l}{\textit{\Benchmark-Text, decision making}} \\
Diagnosis & Predict the next diagnoses of a hospital admission or ED visit, as ICD titles or CCS categories \\
Medication & Predict the next administered medication, prescriptions (as drug names or ATC classes), ED medication reconciliation (as drug names or ATC classes), ED dispensing-cabinet (Pyxis) items, and ICU input and ingredient items \\
Procedure & Predict the next procedures, as ICD titles or CCS categories, and the next ICU procedure items \\
Laboratory \& microbiology & Predict the next laboratory tests and microbiology tests \\
Vitals \& measurements & Predict the next ICU charted items, datetime items, and output items, and the next outpatient measurements (e.g., blood pressure, weight) \\
Imaging report & Predict the next radiology examination \\
Orders & Predict the type of the next provider order \\
Encounter \& flow & Predict the type of the next admission, the next clinical service, and the care unit of the next transfer \\
Next event & Predict which EHR table the next event is recorded in \\
\multicolumn{2}{l}{\textit{\Benchmark-MM}} \\
CXR finding presence & Answer whether a finding is present in a chest X-ray study (EHRXQA) \\
CXR finding enumeration & List the findings of a chest X-ray study (EHRXQA and MedMod) \\
CXR change comparison & Answer how findings change between two chest X-ray studies of the patient (EHRXQA) \\
Clinical deterioration & Predict whether an ICU patient dies within the next 24 hours (MedMod) \\
Mortality & Predict whether an ICU patient dies during the current ICU stay (MedMod) \\
Phenotype & Predict the set of acute and chronic conditions of an ICU stay (MedMod) \\
\bottomrule
\end{tabularx}
\end{table}

\section{Multi-Run Consistency}
\label{app:multirun}
All main-text results use one run per question.
To quantify run-to-run variation, we repeat both inference modes for GPT-5.6 Sol four additional times on \Benchmark-Text and \Benchmark-MM with identical settings, giving five independent runs per mode and benchmark.
\Cref{tab:multirun} lists the overall F1 of each run: the standard deviation across runs is at most 0.4 points for either mode, \Method{} exceeds \Baseline{} in all ten runs, and the paired 95\% confidence interval of the run-averaged difference excludes zero on both benchmarks.
\Cref{tab:multirun_metrics} reports the agreement between runs at the question level with the metrics defined below.
\input{tables/table_multirun}
\input{tables/table_multirun_metrics}

\subsection{Consistency Metrics}
\label{app:consistency_metrics}
Let $s_{ik}\in[0,1]$ be the F1 of question $i$ in run $k$, with $n$ questions and $m{=}5$ runs per mode and benchmark.
\begin{itemize}
\item \textbf{Identical F1.} The share of questions whose F1 is the same in all $m$ runs, $\frac{1}{n}\sum_i \mathbb{1}[s_{i1}=\dots=s_{im}]$.
\item \textbf{Pairwise agreement.} For each of the $\binom{m}{2}$ pairs of runs $(k,l)$, the share of questions with $s_{ik}=s_{il}$, averaged over pairs.
\item \textbf{Krippendorff's $\alpha$.} Treating the runs as $m$ raters of each question, $\alpha = 1 - D_o/D_e$ with the interval metric, where $D_o = \frac{1}{n\binom{m}{2}}\sum_i\sum_{k<l}(s_{ik}-s_{il})^2$ is the observed disagreement within questions and $D_e$ is the same quantity over all $\binom{nm}{2}$ pairs of scores regardless of question. $\alpha=1$ means perfect agreement and $\alpha=0$ agreement at chance level.
\item \textbf{Identical answer.} The share of questions whose final answer set (after the normalization used for scoring) is the same in all $m$ runs. On \Benchmark-MM this is lower than the identical-F1 share because multi-label answers are free text whose wording varies between runs even when the mapped labels, and hence F1, do not.
\item \textbf{Run-averaged paired CI.} Each question's score is first averaged over the $m$ runs of a mode, $\bar{s}_i = \frac{1}{m}\sum_k s_{ik}$; the difference $\bar{s}^{\,\mathrm{agent}}_i-\bar{s}^{\,\mathrm{curated}}_i$ is then resampled over questions (4,000 bootstrap samples) to obtain the 95\% interval reported in \cref{tab:multirun}.
\end{itemize}

\section{Gain by Record Length}
\label{app:length}
\Cref{tab:length_strata} complements \cref{fig:length_strata}: it stratifies the \Benchmark-Text questions by the size of the patient's full pre-cutoff record (in tokens) and reports the difference between \Method{} and Full-Record Input, and between \Method{} and \Baseline{}, within each stratum.
Records in the two longest strata do not fit in the 100k-token context window of Full-Record Input and are truncated oldest-first.
For both models the margin of \Method{} over Full-Record Input increases with record length, from about zero below 30k tokens to +2.1 (Opus 5) and +1.3 (Sol) above 250k tokens, and the margin over \Baseline{} is significant in the longest stratum (+4.6 and +5.3).
\input{tables/table_length_strata}

\section{Case Study: \Method{} vs.\ Full-Record Input}
\label{app:case_full_record}
\begin{figure}[ht]
\centering
\includegraphics[width=\linewidth]{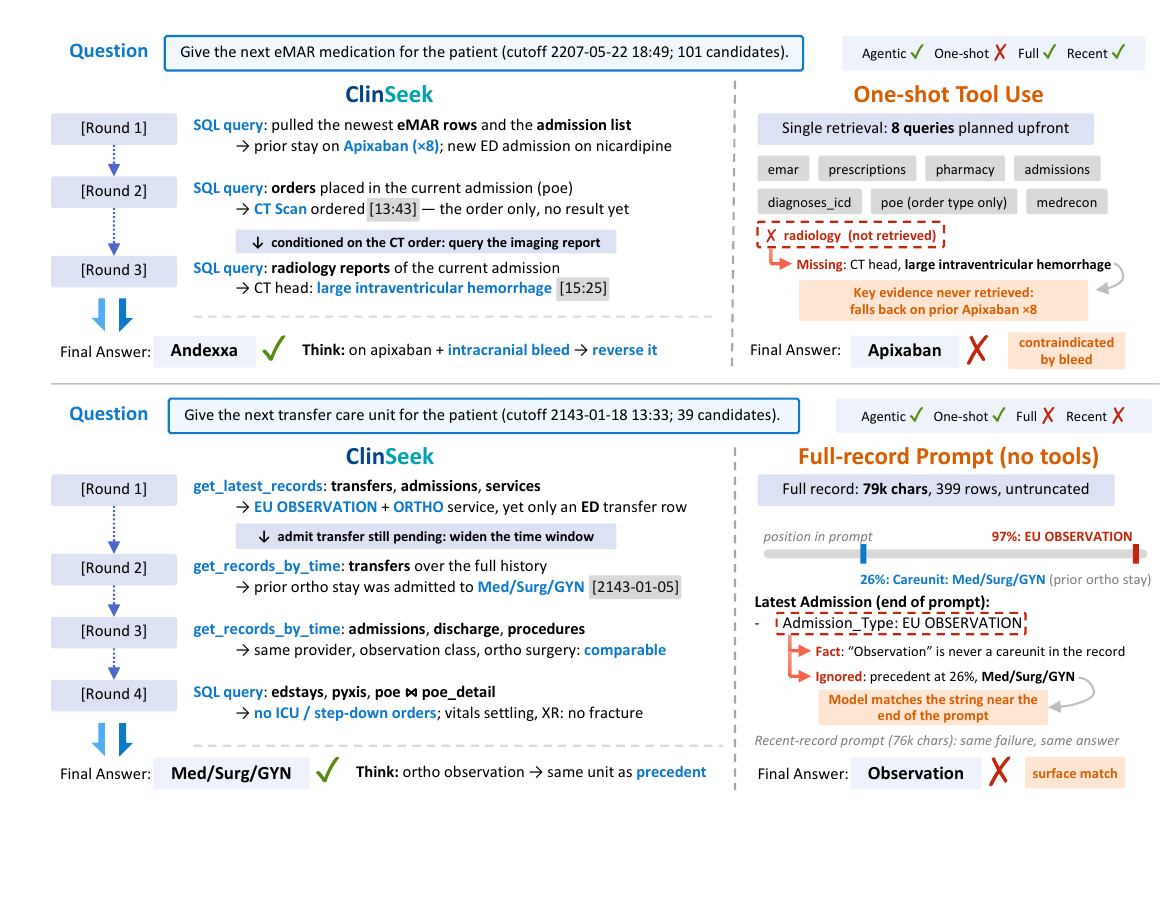}
\caption{\textbf{Case study: \Method{} vs.\ Full-Record Input} (GPT-5.6 Sol, next transfer care unit). Left: the rounds of \Method{} and what each returned; right: the prompt of Full-Record Input.}
\label{fig:case_full_record}
\end{figure}

\Cref{fig:case_full_record} shows the case discussed in \cref{sec:gain_source}, in which the deciding row is present in the Full-Record prompt but the prompt-based model does not use it.

\section{ClinSeekAgent Tool Space}
\label{app:tool_schema}

ClinSeekAgent provides a unified tool interface for multi-source clinical evidence seeking.
The tool space contains EHR tools for patient-specific longitudinal retrieval and image tools for extracting visual evidence from medical images.
\Cref{tab:clinseek_tools_appendix} summarizes the tool names and their functions.
\begin{table}[ht]
\centering
\footnotesize
\setlength{\tabcolsep}{4pt}
\renewcommand{\arraystretch}{1.15}
\caption{
\textbf{ClinSeekAgent tool space.}
EHR tools retrieve patient-specific evidence and image tools extract visual evidence from chest radiographs; \Benchmark-Text uses the 16 EHR tools and \Benchmark-MM all 22 tools.
}
\label{tab:clinseek_tools_appendix}
\begin{tabular}{p{0.07\textwidth} p{0.37\textwidth} p{0.48\textwidth}}
\toprule
\textbf{Source} & \textbf{Tool} & \textbf{Function} \\
\midrule
EHR & \texttt{ehr.load\_ehr} & Load the patient-specific EHR database at the prediction time; required first call. \\
EHR & \texttt{ehr.get\_table\_names} & List the available EHR tables and candidate (vocabulary) tables. \\
EHR & \texttt{ehr.get\_column\_names} & List the columns of a table. \\
EHR & \texttt{ehr.get\_table\_description} & Return the description and column information of a table. \\
EHR & \texttt{ehr.get\_unique\_values} & List the unique values of a categorical column. \\
EHR & \texttt{ehr.get\_records\_by\_time} & Retrieve the records of a table within a time range. \\
EHR & \texttt{ehr.get\_event\_counts\_by\_time} & Count the events of every table within a time range. \\
EHR & \texttt{ehr.get\_latest\_records} & Return all records charted at the latest timestamp of a table. \\
EHR & \texttt{ehr.get\_records\_by\_keyword} & Search the text columns of a table for a keyword. \\
EHR & \texttt{ehr.get\_records\_by\_value} & Retrieve the records whose column exactly matches a value. \\
EHR & \texttt{ehr.run\_sql\_query} & Execute a SQL query over the patient's tables (filtering, joins, aggregation, trends). \\
EHR & \makecell[tl]{\texttt{ehr.get\_candidates\_by}\\\phantom{\texttt{ehr.}}\texttt{\_keyword}} & Search a candidate table for a keyword. \\
EHR & \makecell[tl]{\texttt{ehr.get\_candidates\_by}\\\phantom{\texttt{ehr.}}\texttt{\_fuzzy\_matching}} & Retrieve candidate terms by fuzzy string matching. \\
EHR & \makecell[tl]{\texttt{ehr.get\_candidates\_by}\\\phantom{\texttt{ehr.}}\texttt{\_semantic\_similarity}} & Retrieve candidate terms by embedding similarity. \\
EHR & \texttt{ehr.think} & Record intermediate reasoning and plan the next action. \\
EHR & \texttt{ehr.finish} & Submit the final answer list. \\
\midrule
Image & \texttt{image.dicom\_processor} & Convert a DICOM file to PNG and extract its metadata. \\
Image & \texttt{image.image\_visualizer} & Render an image for inspection. \\
Image & \texttt{image.chest\_xray\_classifier} & Predict probabilities for 18 chest X-ray pathologies. \\
Image & \makecell[tl]{\texttt{image.chest\_xray}\\\phantom{\texttt{image.}}\texttt{\_report\_generator}} & Generate a structured chest X-ray report (findings and impression). \\
Image & \texttt{image.xray\_phrase\_grounding} & Localize a specified finding in a frontal chest X-ray. \\
Image & \makecell[tl]{\texttt{image.chest\_xray}\\\phantom{\texttt{image.}}\texttt{\_segmentation}} & Segment anatomical structures in a chest radiograph. \\
\bottomrule
\end{tabular}
\end{table}

\section{Temporal Policy}
\label{app:temporal_policy}

Each question carries a prediction time $t$.
Because the patient database contains the complete longitudinal record, we construct a question-specific view before answering each question and expose only that view to the model in both inference modes.
The view is constructed table by table according to the following rules.

\begin{enumerate}

\item \textbf{Availability time.}
Each record is exposed according to the time at which its content would have become available in the EHR, rather than according to the clinical event date alone.
A row is retained only if its availability time is before the prediction cutoff $t$.
For structured tables, availability is determined by the corresponding charting or entry timestamp.
For fields that are populated or updated after a row is created, values that would not yet have been available at $t$ are withheld; retaining a row therefore does not imply that all of its fields are visible.
For clinical notes, the document must have been charted and stored by $t$.
For radiology reports, the report must have been finalized by $t$.
Documents completed after the cutoff are excluded even if they describe an earlier encounter, procedure, or image.

\item \textbf{Diagnosis and procedure codes.}
In MIMIC-IV, diagnosis and procedure codes are assigned when an encounter is completed and do not carry a separate time of entry.
We therefore treat the codes of an encounter as available only 24 hours after its discharge or ED departure.
For tasks that predict diagnosis or procedure codes, we additionally hide all such codes from the encounter being predicted, regardless of $t$.

\item \textbf{Readmission tasks.}
These tasks predict whether the patient returns after the index admission, and the prediction cutoff is the discharge time of the index admission.
The discharge summary is included only if it had been charted and stored by the cutoff; summaries completed after the cutoff are excluded.
Other clinical notes follow the same document-availability rule described above.

\item \textbf{Multimodal questions.}
Image and report availability are handled separately.
A chest radiograph linked to the question is visible to the image tools only if it was acquired before the prediction cutoff $t$.
Its associated radiology report is exposed only if the report itself had been finalized by $t$.
Thus, availability of a chest radiograph does not imply availability of its report: reports completed after the cutoff are withheld even when the corresponding image was already available.
The EHR tables follow the same temporal availability rules above.

\end{enumerate}

All EHR retrieval tools, including \texttt{ehr.run\_sql\_query}, operate only on the same question-specific filtered view, so the model cannot retrieve rows or fields excluded by the temporal policy.
The \Baseline prompts are rendered from this same filtered view, ensuring that both inference modes are subject to the same temporal availability constraints; they differ in how the available evidence is accessed and surfaced to the model.
\Cref{app:source_temporal_leakage} compares this policy with the temporal settings of the source benchmarks.

\section{Comparison with the Temporal Settings of the Source Benchmarks}
\label{app:source_temporal_leakage}

\Benchmark-Text is built on EHR-Bench~\citep{liao2025ehr}, and ClinSeek-35B-A3B is evaluated on AgentEHR-Bench~\citep{liao2026agentehr}.
Both benchmarks select patient records by the time of the clinical event.
In MIMIC-IV, some information is recorded later than the event it describes.
For example, a lab result becomes available when it is stored, not when the specimen is collected.
Because \Method{} can query every row and field of the raw record, we compared the settings of the two benchmarks with the time at which each piece of information becomes available.
We then adopt a stricter setting based on availability time (\cref{app:temporal_policy}).
This section lists the main differences between the two settings.

\paragraph{AgentEHR-Bench.}
AgentEHR-Bench filters each table by its first time column and keeps the selected rows in full.
In MIMIC-IV, a single row can combine an event time with fields completed later.
For example, an \texttt{admissions} row selected by \texttt{admittime} also contains the discharge time, discharge location, and in-hospital death flag, and the same table holds a copy of the discharge summary text.
Rows of \texttt{labevents} are selected by \texttt{charttime}, the specimen collection time, whereas results become available at \texttt{storetime}.
Tables without a time column are kept in full.
In our setting, discharge and death fields are withheld until discharge, the copied summary text is removed, lab results are filtered by \texttt{storetime}, and tables without a time column keep only the rows linked to visible parent records, such as visible orders and medication administrations.

\paragraph{EHR-Bench.}
EHR-Bench also orders and selects records by event time.
Discharge summaries are selected by \texttt{charttime} rather than \texttt{storetime}, so a summary stored after the prediction time can appear in the input.
Adjacent \texttt{chartevents} records are merged into groups, and each group is selected by its start time and shown in full.
A group that starts inside the observation window can therefore include measurements taken after the window ends.
Continuous treatments are also selected by start time and can show their final duration or total amount before they end.
In our setting, each row is filtered by its availability time before the \Baseline{} prompt is rendered, and the duration and amount of a treatment are withheld until the treatment is completed.

\paragraph{Examples.}
The following two questions receive different inputs under the two settings.
The first is an AgentEHR-Bench diagnosis prediction question (\texttt{subject\_id=16081909}) with prediction time \texttt{2129-03-07 17:47}.
The discharge summary of this admission has a chart time of midnight on the same day and a store time of 18:19.
Under the original setting the summary is visible at the prediction time, and it states the discharge diagnosis, a Crohn's flare.
The second is an EHR-Bench in-hospital mortality question (\texttt{subject\_id=13001645}, \texttt{idx=5633}) with prediction time \texttt{2187-02-09 16:18}, 24 hours after admission.
The patient died at 21:56 on the same day.
Under the original setting the input contains a discharge summary with a chart time of midnight and a store time of \texttt{2187-02-10 08:19}, and the summary records the death.
Our setting withholds both summaries because they were stored after the prediction time.

All results in this paper are obtained under our setting, so every model and inference mode is compared on the same available information.

\section{Evaluation and Inference Settings}
\label{app:eval_details}

\subsection{Metric}
\label{app:metric}
Let $\hat{Y}$ be the set of normalized labels the model returns for a question and $Y$ the gold set.
We compute $P = |\hat{Y}\cap Y|/|\hat{Y}|$, $R = |\hat{Y}\cap Y|/|Y|$, and $\mathrm{F1} = 2PR/(P+R)$ (0 when $\hat{Y}$ is empty), and average F1 over the questions of a task; task-group and overall scores average over all questions in the group or benchmark.
When $|Y| = 1$ and the model returns one answer, F1 is 1 if the answer matches and 0 otherwise, i.e., accuracy; this covers all risk-prediction tasks and the single-label multimodal questions.

On \Benchmark-Text, labels are matched after whitespace and case normalization; tasks with a fixed candidate list (e.g., admission type, care unit) require the answer to be one of the candidates, and medication tasks with ATC labels are matched on the ATC name.

On \Benchmark-MM, single-label questions are judged by Claude Sonnet 4.6 for semantic agreement with the gold label (polarity for yes/no questions, value for numeric questions, and name for label questions); multi-label questions have each predicted item mapped to the task's gold vocabulary by sentence-embedding cosine similarity (\texttt{all-MiniLM-L6-v2}, threshold 0.55) before set F1 is computed.
Re-judging an unchanged run moves its overall score by about 0.4 points.
All comparisons between settings are paired on questions: we report the mean per-question difference $\Delta$ with a paired-bootstrap 95\% confidence interval (4,000 resamples) and mark $\Delta$ in bold when the interval excludes zero.

\subsection{Inference Settings}
All models are accessed through one OpenAI-compatible inference gateway in front of the providers' deployments; ClinSeek-35B-A3B and its base model are served locally with vLLM behind the same interface.
\cref{tab:eval_hyperparams} lists the settings shared across models and inference modes.
Reasoning is enabled for every model: extended thinking with a 4,096-token budget for Claude models, high reasoning effort for GPT-5.6, and the native reasoning mode of GLM-5.2 and Kimi K2.6.
The Claude and GPT-5.6 APIs do not accept a sampling temperature when reasoning is enabled; temperature 0 is used wherever it is accepted.
Every question is run once.

\paragraph{\Method.}
The agent sees the task instruction, the patient identifier, and the prediction time, and interacts with 16 EHR tools (schema exploration, time-, keyword-, and value-based retrieval, SQL, candidate lookup, \texttt{think}, and \texttt{finish}); multimodal questions add six image tools (visualizer, DICOM processor, CXR classifier, report generator, phrase grounding, segmentation), for 22 tools in total.
Each tool result is truncated to 100,000 characters, and a question ends when the agent calls \texttt{finish} or after 200 rounds.
GLM-5.2 and Kimi K2.6 frequently exhaust the output budget inside their reasoning trace before emitting an answer, in both settings.
For these two models we append one answer-only round to every such question: the conversation so far is replayed and the model is asked to call \texttt{finish} from the evidence already gathered, with no further retrieval.

\paragraph{\Baseline.}
The evidence package of the source benchmark is re-rendered from the policy-filtered record into a single prompt inside a 100k-token window.
Multimodal prompts additionally attach up to four CXR images (longest edge 1,568 pixels) and the linked report text (up to 4,000 characters).
The model is asked to end its reply with an \texttt{<answer>} block, which is parsed into the same answer format as a \texttt{finish} call so that both settings are scored by the same code.

\begin{table}[t]
\centering
\caption{
\textbf{Inference settings.}
The same settings are used for every model and evaluation setting whenever the backend supports them.
}
\label{tab:eval_hyperparams}
\resizebox{0.9\linewidth}{!}{
\begin{tabular}{ll}
\toprule
\textbf{Setting} & \textbf{Value} \\
\midrule
Reasoning & extended thinking, 4,096-token budget (Claude); reasoning effort \texttt{high} (GPT-5.6); native (GLM-5.2, Kimi K2.6) \\
Temperature & 0 where accepted; not settable with reasoning on Claude and GPT-5.6 \\
Maximum output tokens per call & 8,192 \\
Maximum agent rounds & 200 \\
Runs per question & 1 \\
Tool set & 16 EHR tools (text); 16 EHR + 6 image tools (multimodal) \\
Maximum tool-result length & 100,000 characters \\
\Baseline{} context window & 100k tokens \\
Images per question (multimodal) & $\le 4$, longest edge 1,568 pixels \\
Stopping criterion & \texttt{finish} tool call or round limit \\
Metric & mean sample-wise F1; paired-bootstrap 95\% CI \\
\bottomrule
\end{tabular}
}
\end{table}

\section{Training Settings for ClinSeek-35B-A3B}
\label{app:sft_config}

\textbf{Training data.}
GPT-5.6 Sol (reasoning effort high) runs \Method{} with the 16 EHR tools and the strict temporal policy on the 3,000 questions of the AgentEHR-Bench training split, which cover six tasks (diagnoses, procedures, laboratory tests, microbiology tests, prescriptions, and transfers).
We keep the 2,947 trajectories that end with an answer and drop those longer than 52K tokens, leaving 2,681, which we split at random into 2,628 for training and 53 for validation.
The evaluation questions of \cref{tab:SFT_performance} come from the test split and do not appear in the training data.
During data generation the teacher is asked to record its reasoning with a \texttt{think} tool before each action; for training, this reasoning is moved into the student's native thinking block of the following turn, so the student reasons in its thinking channel instead of calling the tool.
The loss is computed on assistant tokens only (reasoning, tool calls, and the final answer).

\textbf{Configuration.}
\Cref{tab:sft_config} lists the training configuration.
We train for three epochs and report the checkpoint at the end of the second epoch, which has the lowest validation loss; validation loss rises in the third epoch.

\begin{table}[ht]
\centering
\small
\setlength{\tabcolsep}{5pt}
\caption{SFT configuration for ClinSeek-35B-A3B.}
\begin{tabularx}{\linewidth}{@{}l >{\raggedright\arraybackslash}X@{}}
\toprule
\textbf{Component} & \textbf{Configuration} \\
\midrule
Base model & Qwen3.5-35B-A3B \\
Teacher model & GPT-5.6 Sol \\
Training objective & Full-parameter supervised fine-tuning on \Method{} trajectories \\
Training data format & Native thinking block and tool-call format \\
Training / validation size & 2,628 / 53 trajectories \\
Maximum sequence length & 52,000 tokens \\
Epochs & 3 (checkpoint after epoch 2 reported) \\
Global / micro batch size & 32 / 1 per GPU \\
Optimizer & Adam ($\beta_1{=}0.9$, $\beta_2{=}0.999$); optimizer states offloaded to CPU \\
Learning rate & $2\times10^{-5}$, cosine decay to $2\times10^{-6}$, 10 warmup steps \\
Weight decay & 0.1 \\
Gradient clipping & 1.0 \\
MoE auxiliary loss & 0.01 \\
MoE z-loss & 0.001 \\
Precision & bfloat16; fp32 optimizer states \\
Backend & verl + Megatron + mbridge \\
Hardware & 2$\times$ H200 GPUs \\
Parallelism & Tensor 2, expert 2, pipeline 1, context 1 \\
Activation recomputation & Full \\
Random seed & 42 \\
\bottomrule
\end{tabularx}
\label{tab:sft_config}
\end{table}

\section{Behavior Analysis of ClinSeek-35B-A3B}
\label{app:sft_behavior}
This section gives the quantitative results behind \cref{sec:sft_behavior}.
We compare the 500 AgentEHR-Bench trajectories of ClinSeek-35B-A3B with those of Qwen3.5-35B-A3B; both models are served in the same way with identical inference settings.
Confidence intervals are obtained by resampling questions with replacement 4,000 times.
The gain in average F1 (+16.1, 95\% CI [14.0, 18.3]) comes from both precision (18.6 to 32.4) and recall (31.1 to 53.3).
Each AgentEHR-Bench task also has a fixed label vocabulary (e.g., CCS diagnosis categories or laboratory test names) in which answers must be expressed; the agent searches it with the candidate lookup tools of \cref{tab:clinseek_tools_appendix} or with SQL, and we call such a search a \textit{label lookup}.

\paragraph{More complex SQL queries.}
We consider only SQL queries over the patient record (471 for the base model, 2,964 for ClinSeek-35B-A3B), not those over the label vocabulary.
A query counts as \textit{complex} if it contains a join, a GROUP BY, an aggregate function (COUNT, MIN, MAX, AVG, SUM), or a comparison on a time column; question-level rates are the fraction of the 500 questions in which the pattern appears at least once (\cref{tab:sft_sql}).
A typical pattern of ClinSeek-35B-A3B first summarizes the orders of the patient and then joins the orders of the current admission with their details:
\begin{footnotesize}
\begin{verbatim}
-- 1) Summarize the patient's orders by type and time span
SELECT order_type, order_subtype, COUNT(*) AS n,
       MIN(ordertime), MAX(ordertime)
FROM poe
GROUP BY order_type, order_subtype
ORDER BY n DESC;

-- 2) Join the orders of the current admission with their details
SELECT p.ordertime, p.order_type, p.order_subtype,
       d.field_name, d.field_value
FROM poe p
LEFT JOIN poe_detail d ON p.poe_id = d.poe_id
WHERE p.hadm_id = 28384443
ORDER BY p.ordertime;
\end{verbatim}
\end{footnotesize}
In contrast, 80\% of the SQL queries of the base model select all columns of a table without further structure.

\begin{table}[ht]
\centering
\small
\caption{\textbf{SQL usage on AgentEHR-Bench.} Question-level rates: fraction of questions in which the pattern appears at least once.}
\label{tab:sft_sql}
\begin{tabular}{lcc}
\toprule
\textbf{Metric} & \textbf{Base} & \textbf{ClinSeek-35B-A3B} \\
\midrule
SQL queries per question & 0.94 & 5.93 \\
Complex queries among all SQL queries & 1.3\% & 23.9\% \\
Questions with a join & 0\% & 41\% \\
Questions with GROUP BY & 0\% & 13\% \\
Questions with an aggregate function & 0\% & 18\% \\
Questions with a time filter & 1\% & 20\% \\
Questions with any complex query & 0.8\% & 53.4\% \\
\bottomrule
\end{tabular}
\end{table}

\paragraph{Reasoning grounded on the patient record.}
\Cref{tab:sft_explore} summarizes how the two models explore the record.
\textit{Tables read} is, per trajectory, the fraction of the patient's available record tables (15.8 on average) that appear in at least one data-reading call, averaged over trajectories; calls that only inspect a table's schema are not counted.
\textit{Targeted queries} is, per trajectory, the fraction of data-reading calls after the first one that look up specific values: a keyword or value lookup, or an SQL query whose WHERE clause filters on more than the patient identifier; trajectories without such later calls are excluded (463 for the base model, 472 for ClinSeek-35B-A3B remain).
The conclusion is unchanged under alternative definitions: excluding SQL over the label vocabulary (12.2\% vs.\ 35.1\%), counting the first data-reading round (10.0\% vs.\ 23.8\%), or also counting time-window reads of at most two days as targeted (22.2\% vs.\ 50.6\%).
\textit{Refers to the current admission} counts the questions whose reasoning text mentions the current admission, encounter, hospitalization, stay, or episode (or ``this admission'', ``index admission'', ``same-day'', ``today'').

\begin{table}[ht]
\centering
\small
\caption{\textbf{How the two models explore the patient record on AgentEHR-Bench.}}
\label{tab:sft_explore}
\begin{tabular}{lcc}
\toprule
\textbf{Metric} & \textbf{Base} & \textbf{ClinSeek-35B-A3B} \\
\midrule
Tables read & 36\% & 72\% \\
Tables read in parallel in the first data-reading round & 1.0 & 5.2 \\
Targeted queries & 13.2\% & 35.8\% \\
Data-reading calls before the first label lookup & 4.4 & 15.7 \\
Questions that read imaging or discharge notes & 12\% & 74\% \\
Questions whose reasoning refers to the current admission & 39\% & 80\% \\
\bottomrule
\end{tabular}
\end{table}

\clearpage

%% file: tables/table_text_appendix.tex
\begin{table*}[ht]
\centering
\small
\caption{\textbf{ClinSeekAgent vs.\ Curated Input on ClinSeek-Bench-Text for more models} (sample-wise F1, \%). Agent = ClinSeekAgent, Curated = Curated Input, $\Delta$ = Agent -- Curated on the same questions; bold marks a gain whose paired-bootstrap 95\% CI excludes zero. Risk Prediction: 18 tasks, $N{=}720$; Decision Making: 27 tasks, $N{=}1080$.}
\label{tab:text_appendix}
\begin{tabularx}{\textwidth}{@{}l *{9}{>{\centering\arraybackslash}X} @{}}
\toprule
\multirow{2}{*}{\textbf{Model}} & \multicolumn{3}{c}{\textbf{Risk Prediction}} & \multicolumn{3}{c}{\textbf{Decision Making}} & \multicolumn{3}{c}{\textbf{Overall}} \\
\cmidrule(lr){2-4} \cmidrule(lr){5-7} \cmidrule(lr){8-10}
& Agent & Curated & $\Delta$ & Agent & Curated & $\Delta$ & Agent & Curated & $\Delta$ \\
\midrule
Claude Opus 4.6 & 81.2 & 81.2 & 0.0 & 40.1 & 43.0 & -2.9 & 56.6 & 58.3 & -1.7 \\
GLM-5.2 & 79.0 & 75.6 & \textbf{+3.4} & 33.5 & 34.6 & -1.1 & 51.7 & 51.0 & +0.7 \\
Kimi K2.6 & 79.6 & 76.4 & +3.2 & 30.1 & 33.5 & -3.3 & 49.9 & 50.6 & -0.7 \\
\bottomrule
\end{tabularx}
\end{table*}

%% file: tables/table_mm_appendix.tex
\begin{table*}[ht]
\centering
\small
\caption{\textbf{ClinSeekAgent vs.\ Curated Input on ClinSeek-Bench-MM for more models} (sample-wise F1, \%; GLM-5.2 was not evaluated on the multimodal benchmark). Categories follow \cref{tab:taxonomy}. $\Delta$ = ClinSeekAgent -- Curated Input; bold marks a gain whose paired-bootstrap 95\% CI excludes zero.}
\label{tab:mm_appendix}
\begin{tabularx}{\textwidth}{@{}ll *{7}{>{\centering\arraybackslash}X} @{}}
\toprule
\textbf{Model} & \textbf{Setting} & \makecell{CXR\\presence} & \makecell{CXR\\enumeration} & \makecell{CXR\\change} & \makecell{Clinical\\deterioration} & \makecell{Mortality} & \makecell{Phenotype} & \makecell{Overall} \\
& & {\scriptsize $N{=}215$} & {\scriptsize $N{=}182$} & {\scriptsize $N{=}222$} & {\scriptsize $N{=}125$} & {\scriptsize $N{=}125$} & {\scriptsize $N{=}120$} & {\scriptsize $N{=}989$} \\
\midrule
\multirow{3}{*}{Claude Opus 4.6} & ClinSeekAgent & 71.9 & 40.5 & 54.2 & 92.0 & 77.6 & 46.8 & 62.4 \\
 & Curated Input & 76.6 & 35.3 & 50.1 & 95.2 & 68.8 & 17.2 & 57.2 \\
 & $\Delta$ & -4.7 & \textbf{+5.2} & +4.1 & -3.2 & +8.8 & \textbf{+29.6} & \textbf{+5.1} \\
\midrule
\multirow{3}{*}{Kimi K2.6} & ClinSeekAgent & 78.3 & 32.8 & 52.0 & 89.6 & 77.6 & 15.8 & 57.8 \\
 & Curated Input & 72.3 & 34.0 & 46.3 & 95.2 & 78.4 & 11.9 & 55.7 \\
 & $\Delta$ & \textbf{+6.0} & -1.2 & \textbf{+5.7} & -5.6 & -0.8 & \textbf{+3.9} & +2.0 \\
\bottomrule
\end{tabularx}
\end{table*}

%% file: tables/table_multirun.tex
\begin{table}[t]
\centering
\small
\setlength{\tabcolsep}{5pt}
\caption{\textbf{Five independent runs of GPT-5.6 Sol under both inference modes} (overall F1, \%). Run 1 is the run reported in \cref{fig:category_gains}. The 95\% CI is the paired-bootstrap interval of the run-averaged per-question difference (\cref{app:consistency_metrics}).}
\label{tab:multirun}
\begin{tabular}{lccc|ccc}
\toprule
& \multicolumn{3}{c|}{\textbf{\Benchmark-Text}} & \multicolumn{3}{c}{\textbf{\Benchmark-MM}} \\
\cmidrule(lr){2-4} \cmidrule(lr){5-7}
& \Method & \Baseline & $\Delta$ & \Method & \Baseline & $\Delta$ \\
\midrule
Run 1 & 59.3 & 57.8 & +1.6 & 62.5 & 60.0 & +2.5 \\
Run 2 & 59.3 & 57.3 & +2.0 & 62.9 & 59.5 & +3.3 \\
Run 3 & 59.6 & 57.4 & +2.2 & 62.6 & 59.9 & +2.7 \\
Run 4 & 59.3 & 57.3 & +2.0 & 61.9 & 59.4 & +2.6 \\
Run 5 & 59.5 & 57.3 & +2.3 & 62.2 & 59.9 & +2.3 \\
\midrule
Mean $\pm$ SD & 59.4$\pm$0.1 & 57.4$\pm$0.2 & +2.0$\pm$0.3 & 62.4$\pm$0.4 & 59.7$\pm$0.3 & +2.7$\pm$0.4 \\
95\% CI of $\Delta$ & & & [0.9, 3.1] & & & [1.5, 4.0] \\
\bottomrule
\end{tabular}
\end{table}

%% file: tables/table_multirun_metrics.tex
\begin{table}[t]
\centering
\small
\caption{\textbf{Question-level agreement between the five runs of GPT-5.6 Sol.} Metrics are defined in \cref{app:consistency_metrics}.}
\label{tab:multirun_metrics}
\begin{tabular}{lcc}
\toprule
\textbf{Metric} & \Method & \Baseline \\
\midrule
\multicolumn{3}{l}{\textit{\Benchmark-Text}} \\
Identical F1 in all five runs (\%) & 80.9 & 84.5 \\
Pairwise identical F1, mean over run pairs (\%) & 90.4 & 92.2 \\
Krippendorff's $\alpha$ of per-question F1 & 0.865 & 0.890 \\
Identical answer set in all five runs (\%) & 70.8 & 74.9 \\
\midrule
\multicolumn{3}{l}{\textit{\Benchmark-MM}} \\
Identical F1 in all five runs (\%) & 70.5 & 73.4 \\
Pairwise identical F1, mean over run pairs (\%) & 79.6 & 82.9 \\
Krippendorff's $\alpha$ of per-question F1 & 0.882 & 0.907 \\
Identical answer set in all five runs (\%) & 29.7 & 41.6 \\
\bottomrule
\end{tabular}
\end{table}

%% file: tables/table_length_strata.tex
\begin{table}[t]
\centering
\small
\caption{\textbf{Gain of ClinSeekAgent by record length on ClinSeek-Bench-Text.} Questions are binned by the token count of the patient's full pre-cutoff record; $\dagger$ marks bins whose records do not fit in the 100k-token context window of Full-Record Input (after reserving room for the model's reasoning and output) and are truncated. Cells give the F1 difference of ClinSeekAgent over the named setting; bold: paired-bootstrap 95\% CI excludes zero.}
\label{tab:length_strata}
\begin{tabular}{llcccc}
\toprule
\textbf{Model} & \textbf{Comparison} & $<$30k & 30--83k & 83--250k$^\dagger$ & $>$250k$^\dagger$ \\
\midrule
\multicolumn{2}{l}{\textit{All questions} ($N$ per bin)} & 467 & 381 & 432 & 520 \\
\cmidrule(lr){1-6}
\multirow{2}{*}{Claude Opus 5} & vs.\ Full-Record Input & 0.0 & +1.5 & +1.6 & +2.1 \\
 & vs.\ Curated Input & 0.0 & +2.0 & +1.7 & \textbf{+4.6} \\
\multirow{2}{*}{GPT-5.6 Sol} & vs.\ Full-Record Input & --1.8 & --0.6 & --0.3 & +1.3 \\
 & vs.\ Curated Input & +0.7 & --1.1 & +0.4 & \textbf{+5.3} \\
\midrule
\multicolumn{2}{l}{\textit{Decision-making questions} ($N$ per bin)} & 320 & 245 & 234 & 281 \\
\cmidrule(lr){1-6}
\multirow{2}{*}{Claude Opus 5} & vs.\ Full-Record Input & +0.9 & +2.3 & +2.6 & +3.6 \\
 & vs.\ Curated Input & +0.3 & +3.9 & +3.6 & \textbf{+8.5} \\
\multirow{2}{*}{GPT-5.6 Sol} & vs.\ Full-Record Input & --1.1 & --1.3 & --2.7 & +3.4 \\
 & vs.\ Curated Input & +1.3 & --1.3 & --0.9 & \textbf{+6.9} \\
\bottomrule
\end{tabular}
\end{table}